\documentclass{article}
\usepackage[letterpaper,textwidth=5.5in,textheight=9in]{geometry}
\usepackage{times}
\usepackage{natbib}

\usepackage[utf8]{inputenc}
\usepackage[T1]{fontenc}
\usepackage{hyperref}
\usepackage{url}
\usepackage{booktabs}
\usepackage{amsfonts}
\usepackage{amsmath}
\usepackage{amssymb}
\usepackage{graphicx}
\usepackage{subcaption}
\usepackage{nicefrac}
\usepackage{microtype}
\usepackage{xcolor}
\usepackage{enumitem}

\title{Corrupted but Correct: Why Vision-Language Models\\ Lie to Themselves Internally}
\author{%
  Arun Josephraj Arokiaraj\textsuperscript{1,2} \qquad
  Chenyi Huang\textsuperscript{1} \qquad
  Zekun Wu\textsuperscript{1,2} \qquad
  Adriano Koshiyama\textsuperscript{2} \\[6pt]
  \textsuperscript{1}Department of Computer Science, University College London\qquad
  \textsuperscript{2}Holistic AI \\[4pt]
  Correspondence: \texttt{arun.arokiaraj.25@ucl.ac.uk}
}
\date{\small Accepted at the VLM4RWD Workshop (Grounded and Faithful Vision-Language Models for Real-World Deployment), NeurIPS 2026}

\begin{document}

\maketitle

\begin{abstract}
A targeted adversarial perturbation can drive a vision-language model's (VLM's) teacher-forced training loss for a fixed target caption to near zero, yet the same model, allowed to generate freely, produces the original, correct description with no trace of the target. We call this dissociation the \emph{train/inference gap}, and give it a precise mechanistic account on Qwen2.5-VL-7B-Instruct using a controlled two-stage PGD attack on 200 held-out COCO images. First, we show that image-level pixel statistics -- including a correctly re-implemented, texture-based attackability measure from the CNN robustness literature -- have essentially no predictive power over which images are corrupted (best predictor $r=-0.050$, $p=0.484$; ridge regression $R^2=0.069$). Second, using the logit lens, we localize the gap to a single autoregressive step: the rank of the target token, conditioned on the correct first token already being generated, is fixed at exactly $3{,}488$ out of $152{,}064$ vocabulary entries for every image and every condition, with zero variance. Third, tracking target-token rank across all 28 LLM decoder layers reveals that the visual encoder corrupts every image's representation by a comparable margin regardless of eventual outcome, but the language model decoder then \emph{differentially arbitrates}: amplifying the corrupted signal for susceptible images and actively suppressing it, past its clean-image baseline, for resistant ones ($p<0.001$, rank-biserial $r=0.579$). A linear probe on the merger hidden state separates these two outcomes with AUC $=0.858$, though we flag and do not paper over a circularity concern in this specific estimate. Together these results argue that adversarial robustness in autoregressive VLMs is substantially a property of the language decoder's prior, not the visual encoder, with direct implications for where faithfulness evaluations and defenses for deployed VLM systems should be targeted.
\end{abstract}

\section{Introduction}
\label{sec:intro}

Vision-language models are increasingly deployed as the perceptual front end of systems that must act on what they see: web-browsing agents, document assistants, and embodied controllers all trust that a model's textual output faithfully reflects the image it was given. A well-studied threat to this trust is the adversarial perturbation -- an imperceptible, $\varepsilon$-bounded change to an image's pixels that flips a model's output toward an attacker-chosen target \citep{szegedy2014intriguing, goodfellow2015explaining, madry2018pgd}. For VLMs this threat is architecturally more complex than for CNN classifiers: a perturbation must survive a visual encoder, a cross-modal merger, and an autoregressive language decoder before it can change a final, open-ended text output, and it has not been clear which of these stages actually determines success or failure.

This paper starts from a specific, reproducible puzzle observed while running such attacks on Qwen2.5-VL-7B-Instruct \citep{bai2025qwen25vl}: projected gradient descent (PGD) can drive the \emph{teacher-forced} training loss for a fixed 3-token target caption to near zero -- meaning the model, if told the correct preceding words, assigns near-total probability to the target -- while the same model, generating freely with no such prompting, produces the original, unperturbed description almost unchanged. Internally, the image is corrupted. Externally, nothing happens. \citet{bailey2024imagehijacks} first noted this qualitatively as a limitation of image-hijack attacks; we treat it as the central object of study and give it a token-level, mechanistic explanation.

Rather than reporting attack success rate as a black-box outcome, we ask three questions about \emph{why} the gap occurs, using 200 held-out COCO val2017 images, a controlled two-stage PGD protocol, and the logit lens \citep{nostalgebraist2020, neo2025interpreting} applied at every layer of both the vision encoder and the language decoder:

\begin{itemize}[leftmargin=1.5em]
    \item \textbf{RQ1 (Encoder susceptibility).} Which images are corrupted, and can this be predicted from pixel-level image statistics alone?
    \item \textbf{RQ2 (Train/inference gap).} At which specific autoregressive step does the attack fail, and is the failure a statistical tendency or an architectural constant?
    \item \textbf{RQ3 (Decoder arbitration).} What does the language model do internally with a corrupted visual signal -- does it passively fail to read a weak signal, or actively suppress it?
\end{itemize}

Our findings motivate a reframing relevant to this workshop's focus on faithful, grounded VLM deployment: adversarial robustness in autoregressive VLMs is not primarily a property of the visual encoder, and evaluating or defending it at the pixel or encoder level alone is likely to miss where the real decision is made.

\textbf{Contributions.} (1) The first demonstration that a CNN-era, texture-based image attackability measure, correctly re-implemented from its original specification, has zero predictive power over VLM adversarial susceptibility. (2) A token-level, architecturally precise localization of the train/inference gap to a single generation step. (3) The first layer-by-layer logit-lens account showing that the LLM decoder -- not the visual encoder -- is the locus of adversarial resistance, amplifying the corrupted signal for susceptible images and suppressing it for resistant ones.

\section{Related Work}
\label{sec:related}

\textbf{Adversarial attacks and the train/inference gap.} PGD \citep{madry2018pgd} and its momentum variant \citep{dong2018mi} are the standard optimizers for $\ell_\infty$-bounded perturbations. \citet{bailey2024imagehijacks} introduce image hijacks and observe, qualitatively, that teacher-forced convergence does not guarantee free-generation success on LLaVA-13B, with the gap widening for longer target sequences; \citet{bengio2015scheduled} formalize the general phenomenon as exposure bias in sequence generation. We quantify this gap at the single-token level on a held-out image set, rather than reporting only aggregate attack success rate.

\textbf{Image attackability.} \citet{raina2023attackability} and \citet{liang2026oti} propose model-free, pixel-level predictors of adversarial susceptibility for CNN classifiers, including an Object Texture Intensity (OTI) measure validated on low-texture, small objects near a CNN decision boundary. \citet{pestana2021defence} extend this with 72 Gray-Level Co-occurrence Matrix texture features and a Random Forest classifier, reaching 75\% accuracy on CNN robustness prediction. Both lines of work are developed and validated exclusively on convolutional image classifiers; we test whether either transfers to an autoregressive VLM's visual encoder.

\textbf{Mechanistic interpretability for VLMs.} The logit lens \citep{nostalgebraist2020} projects an intermediate hidden state through the model's unembedding matrix to read off its ``current'' vocabulary prediction. \citet{neo2025interpreting} validate this tool at visual token positions on \emph{clean} images for LLaVA-1.5 and Qwen2-VL, showing that object information becomes interpretable in vocabulary space by the middle-to-late layers. Concurrently, \citet{nooralahzadeh2026arbitration} study an ``Encoding-Grounding Dissociation'' on clean visual-linguistic conflicts using causal activation patching. We apply the logit lens to \emph{adversarially perturbed} visual tokens for the first time, correlationally, at both the encoder and decoder levels, and connect the resulting layer-wise signal directly to attack outcome.

\textbf{Probing classifiers and language priors.} Linear probes \citep{alain2017probing, belinkov2022probing} test whether a property of interest is linearly encoded in a model's hidden state. \citet{geva2021ffn} show that FFN sublayers act as key-value memories injecting language-prior information into the residual stream, motivating our hypothesis that late-layer FFN-mediated computation, rather than the visual pathway, is responsible for suppressing a corrupted visual signal.

\section{Experimental Setup}
\label{sec:setup}

\textbf{Model.} All experiments use Qwen2.5-VL-7B-Instruct \citep{bai2025qwen25vl} in \texttt{bfloat16}, frozen throughout. Its visual encoder has 32 ViT blocks ($d_{\text{vit}}=1280$, patch size 14, full global attention only at blocks 7, 15, 23, 31; windowed attention elsewhere), followed by a $2{\times}2$-merge projector reducing 1{,}024 patch tokens to 256 merged visual tokens in the LLM's embedding space ($d_{\text{llm}}=3584$), and a 28-layer LLM decoder with vocabulary size $152{,}064$.

\textbf{Dataset.} 200 images drawn from COCO val2017 \citep{lin2014coco}, stratified by supercategory, resized to $448{\times}448$. The target caption is the 3-token sequence \texttt{``A coconut.''} (token IDs 32, 33457, 13), chosen for its semantic distance from every COCO category so that any appearance of the target concept is unambiguous.

\textbf{Attack protocol.} We use a two-stage PGD attack with momentum ($\mu=0.9$), $\varepsilon=16/255$, $\alpha=4/255$, 500 iterations per stage. Stage~1 optimizes the perturbation against the visual encoder alone, with no text prompt, isolating whether the encoder's representation can be pulled toward the target concept independent of any language context. Stage~2 refines the perturbation under the full captioning context with the target caption teacher-forced. Each image is assigned to \textbf{Group A} (susceptible: Stage~1 loss $<0.5$ and the target keyword appears in free-generation output), \textbf{Group B} (train/inference gap: Stage~1 loss $<0.5$ but no keyword hit), or \textbf{Group C} (resistant: Stage~1 loss $\geq 0.5$), giving $n=38$, $9$, and $153$ respectively.

\textbf{Instrumentation.} Forward hooks are registered on all 32 ViT blocks and all 28 LLM decoder blocks. The logit lens is applied at the merger output, at all 56 visual token positions across every LLM layer, and at the final generation position, letting us track the target token's rank in the full vocabulary distribution as it propagates through the model.

\section{RQ1: Encoder Susceptibility Is Not Predictable from Pixels}
\label{sec:rq1}

\begin{table}[t]
\centering
\caption{Pearson correlation of image-level features with Stage-1 loss ($n=200$). None reaches practically meaningful strength; ridge regression on all nine base features gives $R^2_{\text{5-fold CV}}=0.069\pm0.060$.}
\label{tab:rq1}
\small
\begin{tabular}{lrrl}
\toprule
\textbf{Feature} & $r$ & $p$ & Significant? \\
\midrule
Correctly-implemented OTI \citep{liang2026oti} & $-0.050$ & 0.484 & no \\
Whole-image texture intensity (no object mask) & $-0.275$ & 0.0001 & yes (weak) \\
Frequency: high band & $-0.299$ & $<0.001$ & yes (weak) \\
Structural regularity & $-0.033$ & 0.641 & no \\
\bottomrule
\end{tabular}
\end{table}

We compute a nine-feature, model-free image fingerprint (texture intensity, frequency-band energy, structural regularity, object size/count, colour entropy, edge density) for all 200 images, with no access to the model. We also re-implement the Object Texture Intensity (OTI) measure \citep{liang2026oti} exactly to its published specification -- COCO polygon segmentation masks, channel-wise Sobel operator, sum normalized by $C{\times}H{\times}W$ -- after finding that a naive bounding-box, single-channel, per-pixel-mean implementation gives a misleadingly larger correlation ($r=-0.218$).

Table~\ref{tab:rq1} summarizes the result: the correctly implemented OTI measure, the strongest model-free attackability predictor reported for CNN classifiers, has essentially zero correlation with Stage-1 loss on Qwen2.5-VL-7B ($r=-0.050$, $p=0.484$). The best individual predictor we find is whole-image (not object-masked) texture intensity, the opposite of what OTI's object-centric theory predicts. Adding 72 GLCM texture features and switching to a Random Forest classifier \emph{degrades} rather than improves prediction (accuracy $69.0\%\to64.0\%$; ridge $R^2$ falls from $+0.069$ to $-0.055$), which we attribute to genuine architectural mismatch: patch-uniform ViT encoders are more shape-biased and less texture-biased than the CNNs these measures were designed and validated for \citep{dosovitskiy2021vit}. Susceptibility instead follows semantic-category structure that no pixel feature captures -- e.g., trucks and trains are disproportionately susceptible (50--67\%) while birds and beds are fully resistant (0\%) despite comparable or higher texture intensity.

This null result is itself informative: it rules out a simple, actionable heuristic (``avoid highly textured or low-texture images'') for either attackers pre-screening a corpus or defenders hardening one, and motivates looking inside the model rather than at its input.

\section{RQ2: The Gap Is an Architectural Constant, Not a Statistical Tendency}
\label{sec:rq2}

\begin{figure}[t]
\centering
\includegraphics[width=0.85\linewidth]{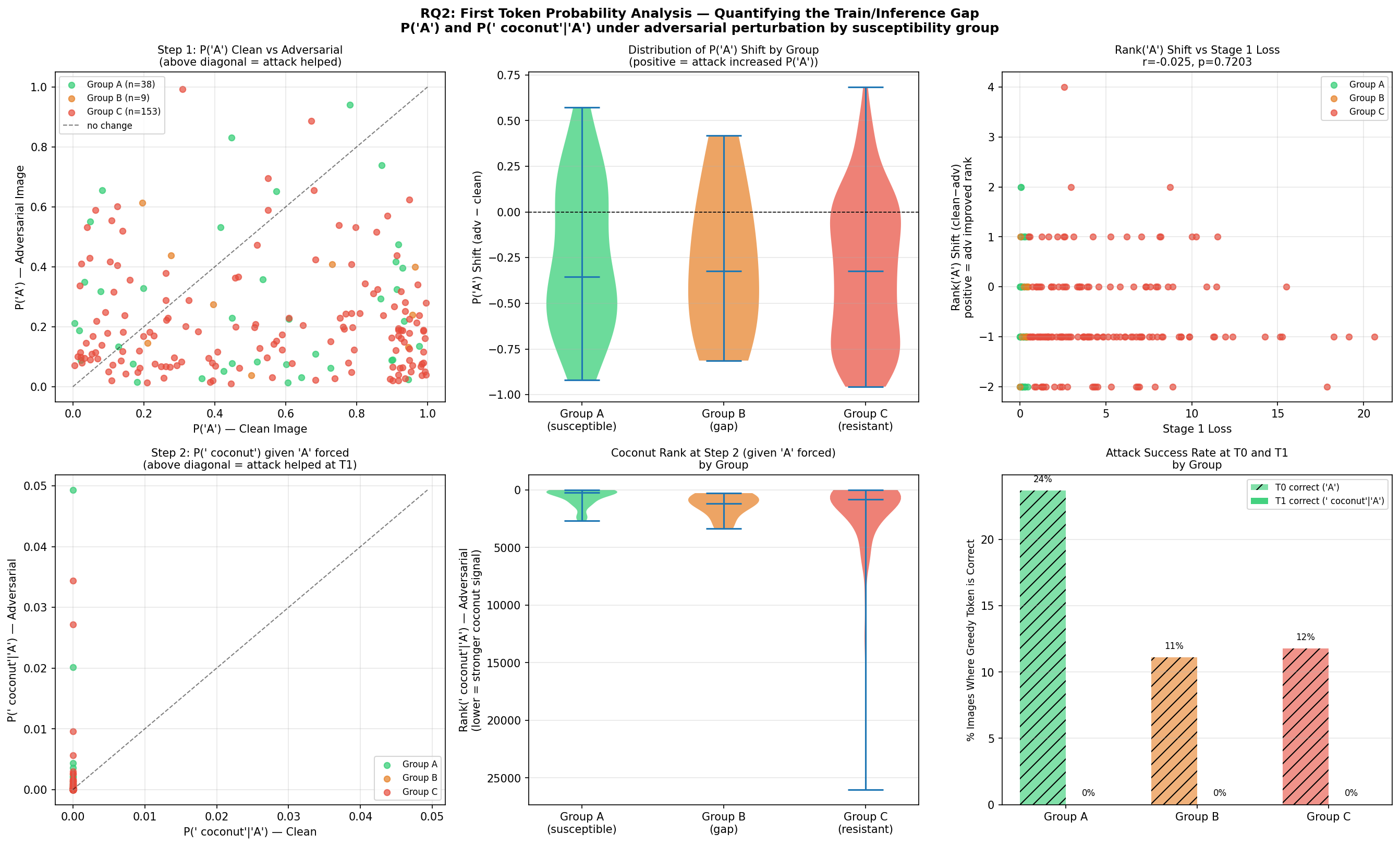}
\caption{Per-token probability analysis across 200 images and all three outcome groups. The rank of the target token \texttt{\_coconut}, conditioned on the correct preceding token \texttt{A} already appearing in context, is a flat line at exactly 3{,}488 with zero variance, for every image, every group, clean and adversarial alike.}
\label{fig:rq2}
\end{figure}

\begin{table}[t]
\centering
\caption{Per-token probability analysis, all 200 images. $\Delta P(\text{A})$: mean change in probability of the correct first token under attack. T0/T1 rate: fraction of images where the model's own greedy choice at that step matches the target.}
\label{tab:rq2}
\small
\begin{tabular}{lrrrrr}
\toprule
\textbf{Group} & $n$ & $\Delta P(\text{A})$ & T0 rate & Rank(\_coconut $\mid$ A) & T1 rate \\
\midrule
A (susceptible) & 38  & $-0.249$ & 23.7\% & 3{,}488 & 0.0\% \\
B (gap)         & 9   & $-0.277$ & 11.1\% & 3{,}488 & 0.0\% \\
C (resistant)   & 153 & $-0.328$ & 11.8\% & 3{,}488 & 0.0\% \\
\bottomrule
\end{tabular}
\end{table}

We record, at each teacher-forced training step, both the log-probability and the full-vocabulary rank of each target token. The attack in fact \emph{decreases} $P(\text{A})$, the correct first token, for every group -- the opposite of the attacker's intent -- meaning teacher-forced loss convergence is not being achieved by making the true first step more likely. The decisive finding is at the second step: the rank of \texttt{\_coconut}, given that \texttt{A} is already forced into context, is exactly $3{,}488$ out of $152{,}064$ vocabulary entries for every one of the 200 images, in every outcome group, under clean and adversarial conditions alike, with zero standard deviation (Table~\ref{tab:rq2}, Figure~\ref{fig:rq2}). Attack success at this step, measured as the model's own greedy choice matching the target, is exactly $0.0\%$ across all 200 images.

This is not a statistical regularity that a stronger or longer attack might move -- it is an architectural constant of this model at this context, invariant to $\varepsilon=16/255$ perturbation. PGD successfully drives down the \emph{average} teacher-forced loss over the three-token target because it can reduce the loss contribution of later tokens even as it fails, entirely and identically, on the second; the average masks a complete, structural failure at a specific step. This gives the train/inference gap a precise mechanistic definition: it is not a matter of degree, but of a single token position at which the perturbation budget cannot move the language prior at all.

\section{RQ3: The Language Decoder, Not the Encoder, Arbitrates the Outcome}
\label{sec:rq3}

\begin{figure}[t]
\centering
\includegraphics[width=0.85\linewidth]{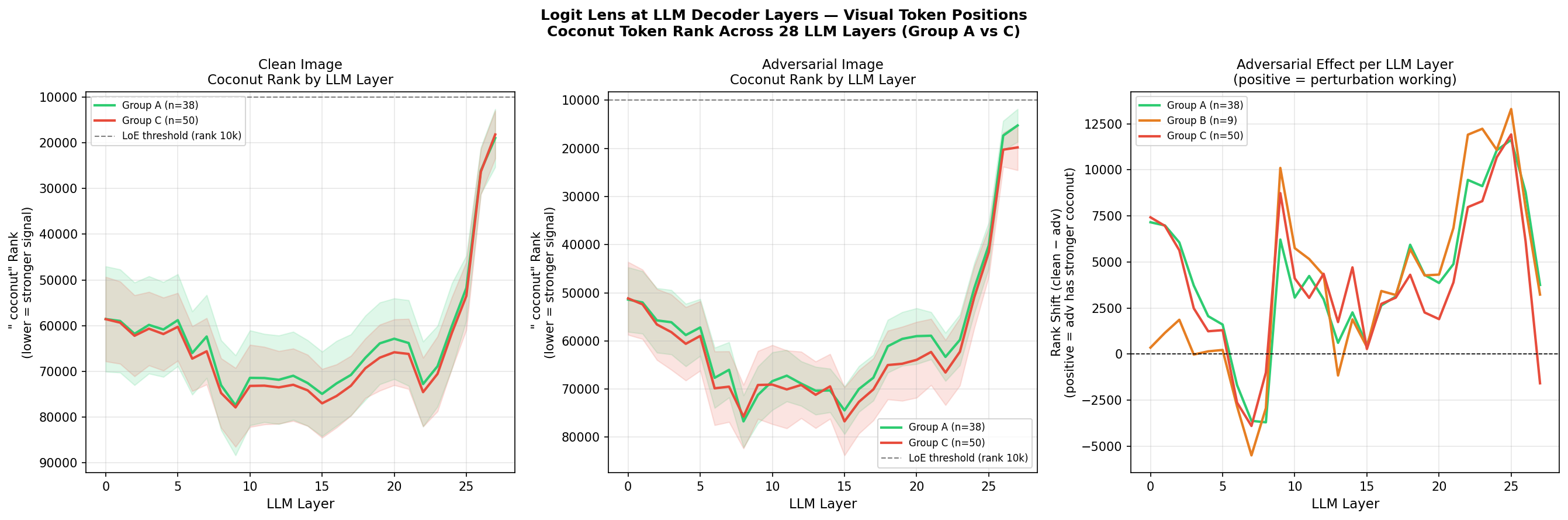}
\caption{Target-token rank shift (adversarial $-$ clean) across all 28 LLM decoder layers, by group. Group A (susceptible) maintains a positive shift toward the target throughout; Group C (resistant) crosses over to \emph{negative} shift -- active suppression, below the clean baseline -- around layers 15--18.}
\label{fig:rq3}
\end{figure}

\begin{table}[t]
\centering
\caption{The two-stage cascade: rank of \texttt{\_coconut} at the vision-language merger output and at the final LLM decoder layer, by group. A negative $\Delta$ means the adversarial rank is lower (closer to the target) than clean; a positive $\Delta$ means the model has moved \emph{further} from the target than its own clean baseline.}
\label{tab:rq3}
\small
\begin{tabular}{lrrrrr}
\toprule
& & \multicolumn{2}{c}{\textbf{Merger output}} & \multicolumn{2}{c}{\textbf{Final LLM layer}} \\
\textbf{Group} & $n$ & Adv.\ rank & $\Delta$ vs.\ clean & Adv.\ rank & $\Delta$ vs.\ clean \\
\midrule
A (susceptible) & 38  & 8{,}714  & $-5{,}724$ & 15{,}224 & $-3{,}754$ \\
B (gap)         & 9   & 10{,}258 & $-2{,}274$ & 17{,}172 & $-3{,}232$ \\
C (resistant)   & 50\textsuperscript{*} & 8{,}571 & $-3{,}619$ & 19{,}787 & $\mathbf{+1{,}580}$ \\
\bottomrule
\end{tabular}
\vspace{2pt}
\\ {\small \textsuperscript{*}Random 50-image subsample of the 153 Group C images for tractable per-layer hook logging.}
\end{table}

Table~\ref{tab:rq3} reveals the cascade. At the merger output -- the first point where the visual representation is expressed in the LLM's embedding space -- the target token's rank drops substantially for \emph{every} group, by a comparable margin (a $-5{,}724$ shift for Group A vs.\ $-3{,}619$ for Group C). The encoder does not distinguish susceptible from resistant images; corruption is essentially uniform at this stage. The divergence appears only downstream, in the LLM decoder: by the final layer, Group A's rank shift has strengthened to $-3{,}754$ (amplification), while Group C's has \emph{reversed sign} to $+1{,}580$ (Mann-Whitney $p<0.001$, rank-biserial $r=0.579$) -- the model has moved the target concept \emph{further away} than it would be for a clean image of comparable content. Figure~\ref{fig:rq3} shows this crossover occurring specifically around layers 15--18 of the 28-layer decoder, consistent with FFN sublayers at this depth injecting corrective language-prior information into the residual stream \citep{geva2021ffn}.

A linear probe trained on the mean-pooled, 3584-dimensional merger hidden state separates Group A from Group C with AUC $=0.858\pm0.076$ on adversarial features, versus $0.630\pm0.034$ on clean features of the same images -- evidence that the eventual outcome is linearly decodable from the model's internal state well before generation begins. \textbf{We flag one limitation explicitly rather than omit it:} the Group A/C labels used to train and evaluate this probe are themselves defined by a threshold on Stage-1 loss, and Stage-1 loss is computed from features of the same representation the probe consumes. Part of the reported separability may therefore reflect this shared dependence rather than an independent representational property; we report the 0.858 figure as suggestive, not conclusive, evidence of linear separability.

Taken together, four independent signals -- pixel statistics (null, \S\ref{sec:rq1}), token-level rank (architectural constant, \S\ref{sec:rq2}), layer-wise logit lens, and the probing classifier -- converge on the same conclusion: adversarial susceptibility is invisible at the input, uniform at the visual encoder, and \emph{decided} at the language decoder.

\section{Discussion: Implications for Grounded and Faithful VLM Deployment}
\label{sec:discussion}

Our results bear directly on the workshop's focus on faithfulness and grounded reasoning in deployed VLM systems. First, \textbf{pixel- or encoder-level defenses are likely to be insufficient}: since the encoder corrupts all images by a comparable margin regardless of eventual outcome (\S\ref{sec:rq3}), a defense operating purely on the visual pathway -- input preprocessing, encoder hardening, or pixel-statistic filtering -- cannot see the signal that actually determines the model's behaviour. A more promising direction, motivated directly by our merger-probe result, is a lightweight anomaly detector placed \emph{after} the merger and \emph{before} the LLM decoder, where the representational difference between eventually-amplified and eventually-suppressed images is largest.

Second, \textbf{the language model's own prior functions as an implicit, but unevenly distributed, defense mechanism.} It reliably overrides adversarial corruption for images with strong, canonical semantic content (Group C: birds, beds), and fails to do so for cluttered or multi-element scenes (Group A: trucks, trains, crowded interiors) that plausibly sit further from any single confident caption. This has a direct, if untested here, implication for agentic and embodied deployment settings, which routinely present exactly this kind of visually diffuse, cluttered observation (a busy webpage, an unstructured tabletop): the implicit protection our results characterize on curated benchmark images may be systematically weaker in precisely the settings this workshop is concerned with.

Third, our results reframe what a faithfulness evaluation for VLM robustness should measure. A protocol that reports only exact-match attack success rate against a fixed target phrase -- as most of the literature we build on does -- would report a 0.0\% success rate here (\S\ref{sec:rq2}) and conclude the model is robust, while missing that the visual encoder is, in fact, substantially corrupted for every image tested. We suggest that faithfulness evaluations for deployed VLMs should measure semantic drift and internal representational corruption directly, not solely whether generation matches an attacker's specific target string.

\section{Limitations}
\label{sec:limitations}

All logit-lens and probing results in \S\ref{sec:rq3} are correlational, not causal; activation-patching experiments intended to establish causal attribution for the layer 15--18 suppression crossover were not technically feasible with our current hook implementation on this architecture and are left to future work. The probing result additionally carries the evaluation-circularity concern noted in \S\ref{sec:rq3}. All findings are demonstrated on a single model, Qwen2.5-VL-7B-Instruct, and a single target caption chosen for semantic distance from COCO categories; we do not claim architecture-general or target-general conclusions in this paper. We deliberately scope this paper to the mechanistic account on 200 images and do not report transfer, ensemble, or agentic-deployment results here, which we treat as a separate line of work with a distinct evaluation methodology.

\section*{Ethics and Responsible Disclosure}
This work characterizes an existing vulnerability class rather than introducing a materially more effective attack; we release the mechanistic analysis to support defense research and do not release optimized attack checkpoints. No experiments involve human subjects or personal data beyond the public COCO val2017 benchmark.

\bibliographystyle{plainnat}
\bibliography{references}

\end{document}